\documentclass{article}

\usepackage{microtype}
\usepackage{booktabs} 
\usepackage{placeins,float,bbm,xspace,empheq,fancybox,amsmath,amssymb,graphicx,epstopdf,epsfig,subfigure,syntonly,times,amsthm}

\usepackage{hyperref}

\usepackage[accepted]{tccmlicml2021}

\icmltitlerunning{Graph Neural Networks for Learning Real-Time
    Prices in Electricity Market}

\DeclareMathOperator*{\argmin}{arg\,min}

\allowdisplaybreaks[3]

\input{mysymbol.sty}

\newcommand{\diag}{\mathrm{diag}}

\begin{document}

\twocolumn[
\icmltitle{Graph Neural Networks for Learning Real-Time
    Prices in Electricity Market}



\icmlsetsymbol{equal}{*}

\begin{icmlauthorlist}
\icmlauthor{Shaohui Liu}{to}
\icmlauthor{Chengyang Wu}{to}
\icmlauthor{Hao Zhu}{to}
\end{icmlauthorlist}

\icmlaffiliation{to}{Department of Electrical and Computer Engineering, University of Texas at Austin, Austin, United States}

\icmlcorrespondingauthor{Shaohui Liu}{shaohui.liu@utexas.edu}
\icmlcorrespondingauthor{Hao Zhu}{haozhu@utexas.edu}

\icmlkeywords{graph neural network, optimal power flow, real-time electricity pricing, transfer learning}

\vskip 0.3in
]



\printAffiliationsAndNotice{}  

\begin{abstract}

Solving the optimal power flow (OPF) problem in real-time electricity market improves the efficiency and reliability in the integration of low-carbon energy resources into the power grids. To address the scalability and adaptivity issues of existing end-to-end OPF learning solutions, we propose a new graph neural network (GNN) framework for predicting the electricity market prices from solving OPFs. The proposed GNN-for-OPF framework innovatively exploits the locality property of prices and introduces physics-aware regularization, while attaining reduced model complexity and fast adaptivity to varying grid topology. Numerical tests 
have validated the learning efficiency and adaptivity improvements of our proposed method over existing approaches. 

\end{abstract}


\section{Introduction}\label{sec:intro}

Electricity market pricing is one of the most crucial tasks of  operating large-scale power grids. As part of the deregulated electricity market,  real-time market determines the incremental adjustment to the day-ahead dispatch by solving the optimal power flow (OPF) problem  \cite{cain2012history}, 
which aims at the most economic decisions for the flexible generation or demand while satisfying a variety of safety-related network constraints. The real-time OPF or market pricing is instrumental for ensuring high efficiency and reliability of grid operations \cite{cain2012history}, particularly under the increasing integration of intermittent and variable resources towards a low-carbon energy future.   
 

The accurate ac-OPF problem is known to incur high computation complexity due to its non-linear, non-convex formulation \cite{cain2012history}. For efficient online solution, machine learning (ML) techniques have been recently advocated through extensive off-line training of neural network (NN) models. Existing ML-for-OPF approaches have focused on identifying the active constraints \cite{misra2018learning,deka2019learning,chen2020learning},  finding a warm start for iterative OPF solutions \cite{baker2019learning}, or addressing the feasibility issue \cite{pan2019deepopf,guha2019machine,zamzam2020learning}.
Almost all of them rely on \textit{end-to-end} NNs, which incur high model and computation complexity for large-scale power grids. In addition to scalability issue, they need to be constantly re-trained whenever the system inputs change as a result of frequently varying grid resources or topology. Thus, existing approaches fall short in efficiently transferring the knowledge obtained from off-line training into fast, adaptive online OPF decisions. 

To tackle these challenges, we propose to leverage the graph neural networks (GNNs) to design a topology-aware OPF learning framework. The GNN architecture \cite{kipf2016semi,gama2020graphs,garg2020generalization} can effectively incorporate graph-based embedding of nodal features and explore the topology structures of the underlying prediction models. While a very recent work \cite{owerko2019optimal} has used GNNs to predict OPF's nodal power injections, the latter mainly depends on the cost of dispatching each resource and does not share any topology-based similarity, or the \textit{locality property} that is ideal for GNN-based predictions. Hence, we instead advocate to predict the actual OPF outputs for electricity market, namely the locational marginal prices (LMPs) known as the real-time market signals \cite{wood2013power}. As LMPs relate to the duality analysis for OPF, their  dependence on grid topology has been recognized in \cite{jia2013impact,geng2016learning}. 

To this end, we have introduced the ac- and dc-OPF problem formulations (Section \ref{sec:ps}) and exploited the topological structure of LMPs to design a GNN-for-OPF learning framework (Section \ref{sec:gnn}). This physics-aware approach not only capitalizes on the locality property of LMPs, but also motivates meaningful regularization on the feasibility of OPF line limits. Numerical results (Section \ref{sec:test}) have demonstrated the high prediction performance of the proposed GNN-for-LMP approach at reduced model complexity, while confirming its topology adaptivity as an effective transfer learning tool to deal with fast varying grid topology in real-time markets.

\section{Real-Time Market Modeling}
\label{sec:ps}


Consider a power grid modeled by an undirected graph $G=(\mathcal{V},\mathcal{E})$. The node set $\mathcal{V}$ consists of $N$ nodes, each connected to loads or generators, while the edge set $\mathcal{E}\in \mathcal{V}\times \mathcal{V}$ includes transmission lines or transformers. 
Let $\bbp, ~\bbq \in \mathbb{R}^N$ collect the nodal active and reactive power injections, respectively; and similarly for voltage $\bbv \in \mathbb{R}^N$. Given the network admittance (Y-bus) matrix $\bbY\in \mathbb{R}^{N\times N}$, 
the ac-OPF problem is formulated as
\begin{subequations}
\label{eq:acopf}
\begin{align}
 \min_{\bbp,\bbq,\bbv}~~  & \textstyle\sum_{i=1}^N c_i(p_i) \\
\mbox{s.t.}\;~ &\bbp + \mathrm{j}\bbq = \diag(\bbv)(\bbY\bbv)^{*} \label{eq:acopf_equal}\\
&\underline{\bbV} \leq |\bbv| \leq \bbarbbV \label{eq:acopf_v}\\
&\underline{\bbp} \leq \bbp \leq \bbarbbp \label{eq:acopf_p}\\
&\underline{\bbq} \leq \bbq \leq \bbarbbq \label{eq:acopf_q}\\
& \underline{f}_{ij} \leq f_{ij}(\bbv) \leq \bbarf_{ij} \,, \,\;\; \forall(i,j) \in \mathcal{E} \label{eq:acopf_f}
\end{align}
\end{subequations}
where $c_i (\cdot)$ is a convex (typically quadratic or piece-wise linear) cost function for flexible nodal injections. 
The equality \eqref{eq:acopf_equal} ensures nodal power balance, while constraints \eqref{eq:acopf_v}-\eqref{eq:acopf_f} list the various operational limits such as line flow limits in \eqref{eq:acopf_f}. This general OPF \eqref{eq:acopf} includes both flexible generation and demand, with negative injections for the latter.  
 
To simplify the nonlinear, non-convex problem \eqref{eq:acopf}, the linear dc-OPF is widely used for solving $\bbp$ only, as
\begin{subequations}
\label{eq:dcopf}
\begin{align}
\textstyle \min_{\bbp} ~~\textstyle & \textstyle \sum_{i=1}^N c_i(p_i) \\
\mbox{s.t.}\;~ &\mathbf{1}^\top \bbp = 0 \label{eq:dcopf_equal}\\
& \underline{\bbp} \leq \bbp \leq \bbarbbp \label{eq:dcopf_p}\\
&  \underline{\bbf}  \leq \bbS\bbp \leq {\bbarbbf} \label{eq:dcopf_f}
\end{align}
\end{subequations}
where matrix $\bbS$ is the injection shift factor (ISF) matrix to form the line flow $\bbf = \bbS \bbp$ with the limit $\underline{\bbf} = -\bbarbbf$. Compared to \eqref{eq:acopf}, the dc-OPF problem omits the modeling of reactive power and voltage, and also uses lossless linearized power flow to simplify power balance as in \eqref{eq:dcopf_equal}. The accuracy of dc-OPF can be improved by considering better linearization around the operating points and including line losses; see e.g., \cite{garcia2019non}. As the resultant constraints are still linear, the generalized dc-OPF problem can be easily computed using off-the-shelf convex solvers. 

Learning the OPF solutions amounts to obtaining the mapping from the uncontrolled problem inputs to the OPF outputs. In real-time market \eqref{eq:acopf}-\eqref{eq:dcopf}, nodal injections have  uncontrollable components $\bbp^u$ and $\bbq^u$ from variable demand or renewable resources. They in turn affect the  limits of respective injections in \eqref{eq:acopf}-\eqref{eq:dcopf}. In addition, the cost function $c_i(\cdot)$ depends on the offers submitted by generation or load serving entities (LSEs), thus varying for each OPF instance as well. 
Hence, for each node $i$ the input variables include  $\bbx_i \triangleq [ \bbarp_i, \underline{p}_i, \bbarq_i, \underline{q}_i, \bbc_i ] \in \mathbb{R}^{d}$
, with $\bbc_i$ denoting the $(d-4)$ parameters used for defining the nodal cost function. For example, quadratic cost is given by the quadratic and linear coefficients, while piece-wise linear one by the change points and gradient of each linear part. Due to increasing variability of resources and offers, the real-time OPF problems may experience dramatic changes from instance to instance. Given this vast variability, it is beneficial to develop a learning-based approach that can enable efficient real-time market operations. 




\section{Topology-aware Learning for Market Prices}
\label{sec:gnn}

We advocate a topology-aware graph neural network (GNN) based framework for learning real-time prices that attains high learning efficiency and topology adaptivity. Before introducing GNNs, we first discuss how locational marginal prices (LMPs), the outputs of OPF, are connected to the grid topology $G$.   LMPs are market signals used by each generator or demand to determine the flexible power injection in order to minimize its own cost. To show the topology dependence, consider the simple convex dc-OPF problem \eqref{eq:dcopf}, for which dual variables $\lambda$, and $[\underbar{\bbmu}$; $\bar{\bbmu}]$ are introduced for constraints \eqref{eq:dcopf_equal} and \eqref{eq:dcopf_f}, respectively, with \eqref{eq:dcopf_p} kept as an implicit constraint. Given the optimal dual variables (denoted by $^*$), the nodal LMP vector 
is given by 
\begin{align}
    \bbpi \triangleq \lambda^* \cdot \mathbf{1} -\bbS^\top (\bar{\bbmu}^* - \underbar{\bbmu}^*)
    \label{eq:lmp_def}
\end{align}
using the ISF matrix $\bbS$. Interestingly, vector $(\bar{\bbmu}^* - \underbar{\bbmu}^*)$ indicates the congested lines due to complimentary slackness \cite{boyd2004convex} ; i.e., $\bbarmu_\ell^*(\underline{\mu}_\ell^*)=0 $ if and only if line $\ell$ is reaching limit $\bbarf_\ell (\underline f_\ell)$. Clearly, the LMP $\bbpi$ only depends on those congested lines that have non-zero $(\bbarmu_\ell^*-\underline{\mu}_\ell^*)$. Interestingly, matrix $\bbS$ strongly depends on the graph topology such that $\bbpi$ has the \textit{locality} property that is perfectly suited for GNNs.  Typically, only a few transmission lines are actually congested \cite{price2011reduced}. Thus, LMPs tend to be similar within the neighboring nodes. Formally, matrix $\bbS$ depends on graph incidence matrix  $\bbA_r$ and a diagonal matrix with line reactance values $\bbX = \diag\{x_{ij}\}$, as well as the resultant weighted graph Laplacian matrix $\bbB_r = \bbA_r^\top \bbX^{-1} \bbA_r$. Both $\bbA_r$ and $\bbB_r$ are reduced from the original matrices by eliminating a reference node to obtain the full-rank counterparts. Given the compact singular value decomposition (SVD) $\bbA_r^\top \bbX^{-\frac{1}{2}} = \bbU\bbSigma\bbV^\top$, we can write the ISF matrix as   
\begin{align}
\bbS^\top =  \bbB_r^{-1} \bbA_r^\top \bbX^{-1} = \bbU \bbSigma^{-1} \bbV^\top \bbX^{-\frac{1}{2}} \label{eq:matS}
\end{align}
with the eigen-decomposition $\bbB_r = \bbU\bbSigma^2\bbU^\top$. Thus, the LMP vector $\bbpi$ in \eqref{eq:lmp_def} is exactly generated by the eigen-space of the Laplacian $\bbB_r$, which can be viewed as a graph shift operator (GSO) \cite{scaglione2021gsp}. 
Accordingly, it strongly depends on the graph topology, which motivates one to use the topology-aware GNN models for prediction. Note that even though this LMP analysis corresponds to the simple dc-OPF, similar intuitions also hold for the ac-OPF problem; see e.g., \cite{garcia2019non}. 

In the OPF problem, we aim to obtain the function $f(\bbX) \rightarrow \bbpi$, where the input $\bbX \in \mathbb R^{N\times d}$ has the nodal features $\{\bbx_i\}$ as its rows. To model $f(\cdot)$ using fully-connected NN (FCNN), the input to first layer $\bbX_0$ can be a vector embedding of $\bbX$, with each layer $t$ as
\begin{align}
    \mathbf{X}_{t+1}=\sigma(\mathbf{W}_t \mathbf{X}_t+\mathbf{b}_t),~~\forall t = 0,\ldots, T-1
\end{align}
where $\bbW_t$ and $\bbb_t$ are parameters to be learned, while $\sigma(\cdot)$ is the nonlinear activation like ReLU. 
Albeit generalizable  to a variety of \textit{end-to-end} learning tasks, the FCNN models would incur significant scalability issue for large-scale OPF learning. 
It is possible to reduce the layer complexity by using the graph topology, leading to graph-pruned NNs. For example, the {\textit{graph-induced deep NN} (GiDNN)} developed in \cite{zamzam2020physics} sparsifies matrix $\bbW_t$ according to the graph topology. 
By pruning out a majority of blocks in $\bbW_t$,  the total number of parameters is reduced.


Inspired by the graph signal viewpoint on $\bbpi$ arising from the structure in \eqref{eq:matS}, we propose to systematically reduce the prediction model complexity by leveraging the GNN architecture \cite{isufi2020edgenets,ma2020deep,kipf2016semi}. As a special case of NNs, GNNs take the input features $\{\bbx_i\}$ defined over graph nodes in $\ccalV$, with each layer aggregating only the neighboring node embeddings. In this sense, it is ideal for predicting output labels having locality property as a result from graph diffusion processes. 
To define the GNN layers, consider again the feature matrix $\bbX$ as the input to the first layer $\bbX_0$, and each layer $t$ now becomes: 
\begin{align} 
\bbX_{t+1} = \sigma \left( \bbW \bbX_{t} \bbH_t + \bbb_t \right),~~\forall t = 0,\ldots, T-1 \label{eq:gnn}
\end{align}
where the feature filters $\{\bbH_t\}$ are the $(d_t \times d_{t+1})$ parameter matrices that are learned through training, which do not change with system size $N$. The key of GNNs lies in the graph convolution filter $\bbW \in \mathbb R^{N\times N}$ such that the node embedding is updated by neighborhood aggregation. Matrix $\bbW$ can be the (weighted) graph Laplacian or adjacency matrix, or its normalized version for stability concerns \cite{isufi2020edgenets}. For better performance, it can also be learned through training, leading to a bi-linear filtering process in \eqref{eq:gnn} as developed in \cite{isufi2020edgenets}. In this case, $\bbW$ has the sparsity structure as the graph Laplacian with number of non-zero parameters scaling with that of edges. Clearly, the GNN architecture can significantly reduce the number of parameters per layer. As the average node degree of real-world power grids is around 2 or 3 \cite{birchfield2016grid}, we have the following result. 
%



\begin{assumption}
The edges are very sparse, and the number of edges $|E| \sim \ccalO(|V|) = \ccalO(N)$. \label{as:E}
\end{assumption}

\begin{proposition}
Under (AS\ref{as:E}) and by defining $D = \max_t\{d_t\}$, the number of parameters for each bi-linear GNN layer in \eqref{eq:gnn} is $O(N + D^2)$.
\label{prop:complexity}
\end{proposition}
\vspace*{-2mm}
This complexity order result follows easily from checking the number of nonzero entries in $\bbW$ and $\bbH_t$ in \eqref{eq:gnn}. Trainable graph filter $\bbW$ only increases the complexity by the number of edges, which scales linearly with $N$ thanks to (AS\ref{as:E}). Compared to $\ccalO(N^2D^2)$ as the number of parameters in each FCNN layer, the GNN  architecture scales very gracefully with the network dimension. Thanks to the locality property of LMP $\bbpi$, our proposed design can greatly improve computation time and generalization performance by utilizing the reduced-complexity GNN models. 



\textbf{Feasibility-based Regularization:} As OPF is a network-constrained problem, we design the loss function for learning LMPs that can account for the solution feasibility and constraints. Note that for dc-OPF problem, the LMP fully determines the decision variables in $\bbp$. Based on the KKT optimality condition \cite{boyd2004convex}, the predicted LMP $\hhatbbpi$ allows to obtain the optimal nodal injection, as
    \begin{align}
    p_i^* = \argmin_{\underline{p}_{i} \leq p_i \leq \bbarp_{i}}~~ &c_i(p_i) - \hat{\pi}_i p_i, ~~\forall i \in \ccalV \label{eq:pi2pg}
    \end{align}
For quadratic (or generally strongly-convex) cost functions, the solution is unique by comparing the unconstrained minimum with the boundary points $[\underline{p}_{i}, \bbarp_{i}]$. As for (piece-wise) linear cost functions, this also holds for most nodes if the derivative $c'_i(p_i^*) \neq  \hat \pi_i$. Otherwise, the optimal $p_i^*$ at the other nodes can still be computed from the power balance of the full system and congested lines. 

Using this result, we advocate the following chain to generate the corresponding nodal injection and line flow solutions to the predicted LMP $\hhatbbpi$:
    \begin{align*}
        \bbX \xrightarrow{\text{$f(\bbX;\bbtheta)$}} \hhatbbpi
        \xrightarrow{\text{\eqref{eq:pi2pg}}} {\hhatbbp}^*(\hhatbbpi) \xrightarrow{\text{$\bbS$}} {\hhatbbf}^* (\hhatbbpi)
\end{align*}
where $f(\cdot;\bbtheta)$ denotes the GNN model with weight parameter $\bbtheta$ according to \eqref{eq:gnn}. Hence, the predicted ${\hhatbbp^*}$ is strictly feasible for \eqref{eq:dcopf_p}, while the predicted $\hhatbbf^*$ can be used to regularize the GNN loss function by enforcing the feasibility of \eqref{eq:dcopf_f}. This way, the loss function for GNN training becomes
\begin{align}
        \ccalL \left( \bbtheta \right) &:= \|{\bbpi} - \hhatbbpi \|^2_2 
    + \lambda \| \sigma(|\hat{\bbf}^* (\hhatbbpi)| - \bbarbbf) \|_1  \label{eq:loss}
\end{align}
where the second term captures the total line flow violation of the limit $\bbarbbf$, leading to LMP prediction more amendable to feasibility. Additional regularization terms on $\hhatbbp$ can be introduced such as the infinity error norm in predicting $\bbpi$.



\section{Numerical Results}

\label{sec:test}


\begin{figure}[tb!]
\centering
\subfigure{\label{fig:a_error}\includegraphics[width=90mm]{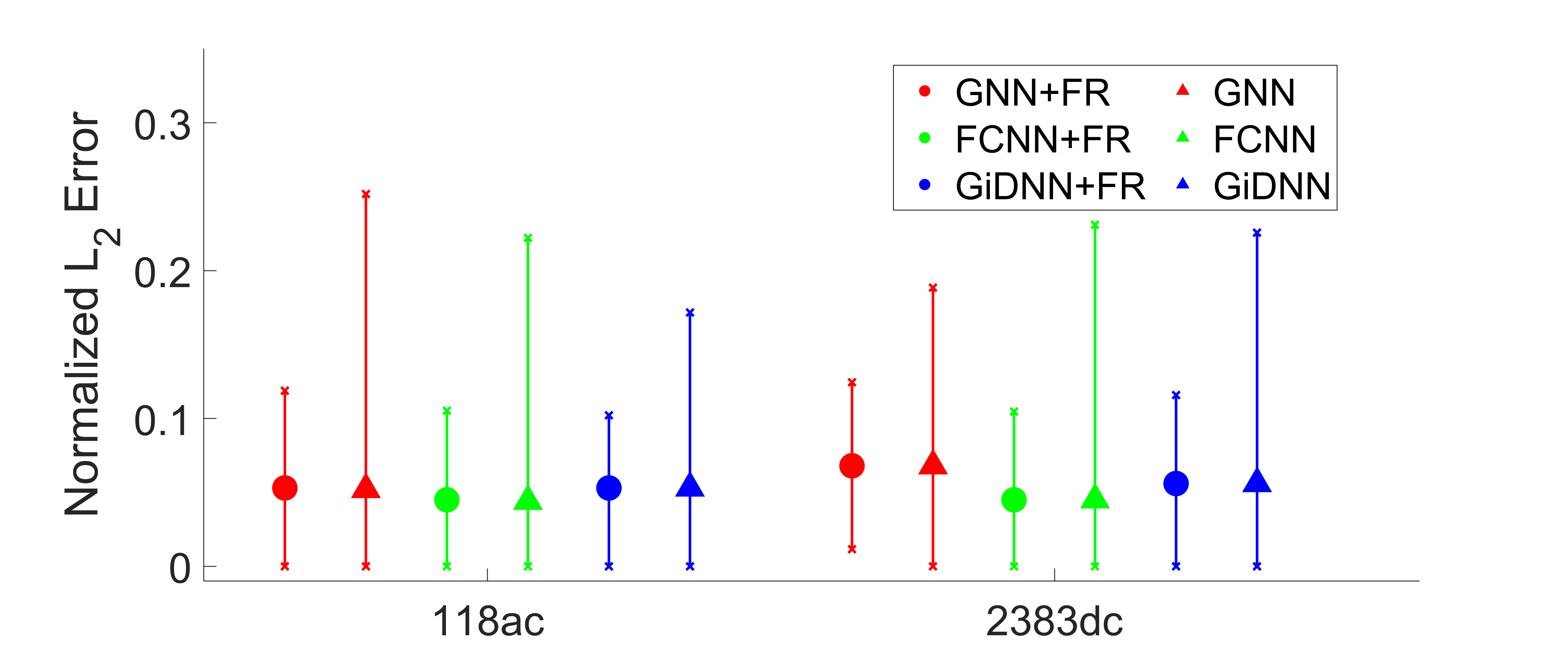} 
\vspace*{-5mm}}
\subfigure{\label{fig:b_feas}\includegraphics[width=90mm]{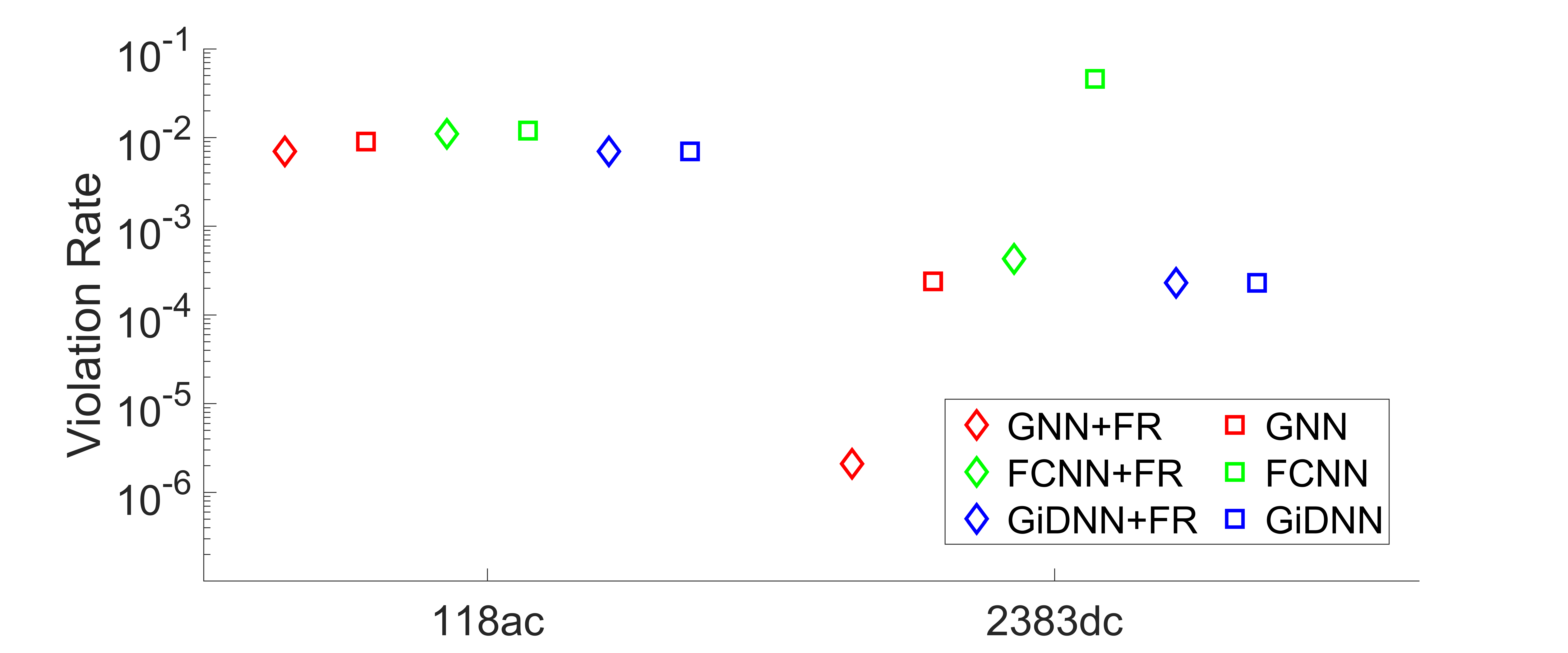}}
\vspace*{-5mm}
\caption{Comparison of GNN, FCNN, and GiDNN models, for both MSE loss and the feasibility regularized (FR) one, in terms of (top) the normalized $L_2$ error in predicting $\bbpi$ (mean $\pm$ standard deviation); and (bottom) the  violation rates of line limits (total violation level versus total limit) for the feasibility performance.}
\vspace*{-5mm}
\label{fig:training_err}
\end{figure}

This section presents the efficiency and scalability results for the proposed GNN-based algorithms by using the 118- and 2383-bus systems from the IEEE PES PGLib-OPF benchmark library \cite{babaeinejadsarookolaee2019power}. A small example on topology adaptivity is also included to demonstrate the proposed GNN models can quickly adapt to varying grid topology in real-time operations. We generated the datasets from solving the ac/dc-OPF problems for each system in MATPOWER \cite{matpower}, by randomly perturbing the operating conditions (limits for $\bbp$/$\bbq$ and the quadratic cost coefficients of $c_i$). GNN models with high-order graph filter \cite{owerko2019optimal}  and \texttt{relu} activation for each layer were implemented by PyTorch library. GNN models, and the benchmark FCNN and GiDNN models (implemented by PyTorch library as well) were tested on Google Colaboratory using the Nvidia Tesla V100 for training acceleration. 

\begin{figure}[tb!]
\centering
\includegraphics[width=75mm]{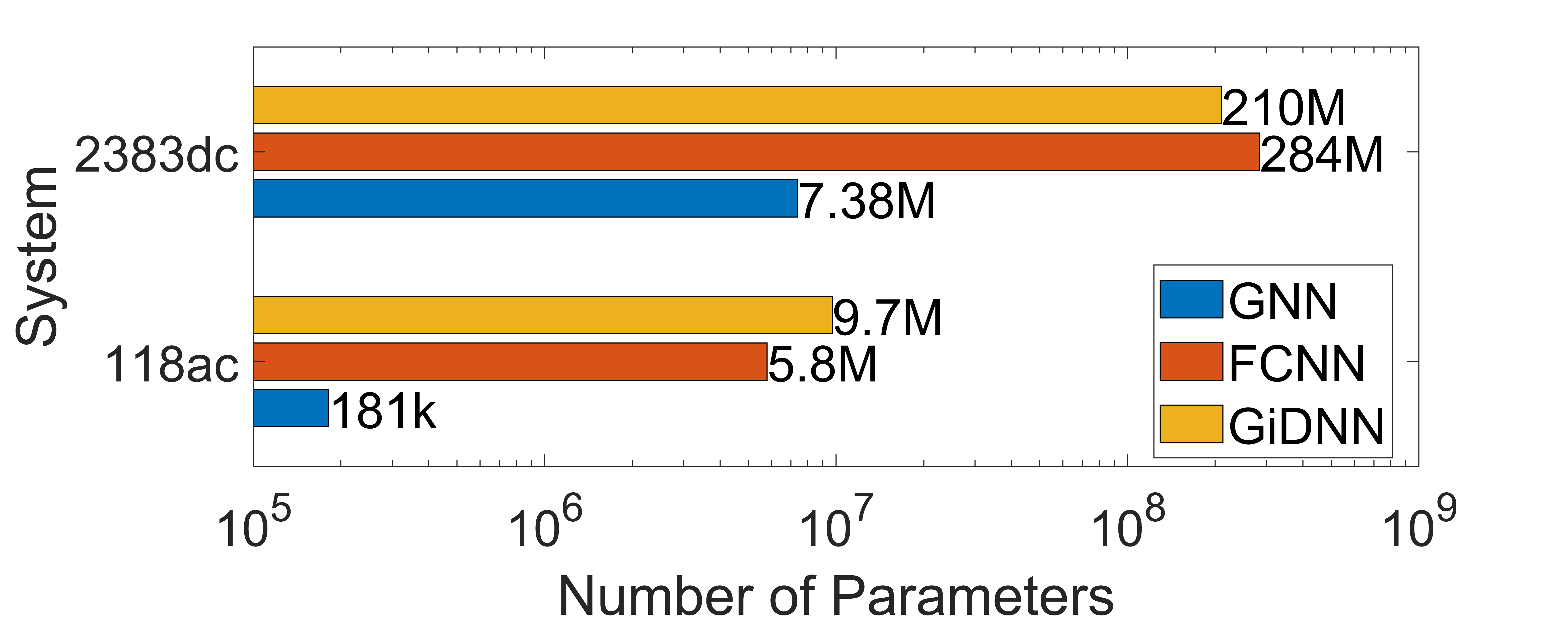}
\vspace*{-4mm}
\caption{The model complexity of GNN, FCNN, and GiDNN in number of parameters of 118ac and 2383dc systems.}
\vspace*{-3mm}
\label{fig:training_time}
\end{figure}

Figure \ref{fig:training_err} compares the performance of proposed GNN-based models with FCNN and GiDNN ones, including those using the feasibility regularized (FR) loss function in \eqref{eq:loss}.  The normalized $L_2$ error in predicting $\bbpi$ and the violation rate of line flow limits are considered, for the ac-OPF of 118-bus system and dc-OPF of 2383-bus system. Clearly, the performance of proposed GNN models is comparable to that of FCNN and GiDNN ones. The FR loss function design has shown to improve the feasibility of OPF predictions for the larger 2383-bus system. In addition, it can accelerate the training process as corroborated by the actual number of epochs for convergence (not included due to page limit). Compared to FCNN models, GNN ones clearly improve the learning accuracy and feasibility in the 2383-bus system prediction. Hence, the proposed GNN architecture along with feasibility based loss function design has shown effective in predicting feasible OPF solutions, especially for large-scale systems. To demonstrate GNN's reduced complexity, Figure \ref{fig:training_time} compares the total number of parameters for each model. The parameter number is indeed reduced by utilizing the topology-based structure of the GNN architecture. 


%

\begin{figure}[tb!]
\centering     
\subfigure[Pre-trained]{\label{fig:a}\includegraphics[width=40.5mm]{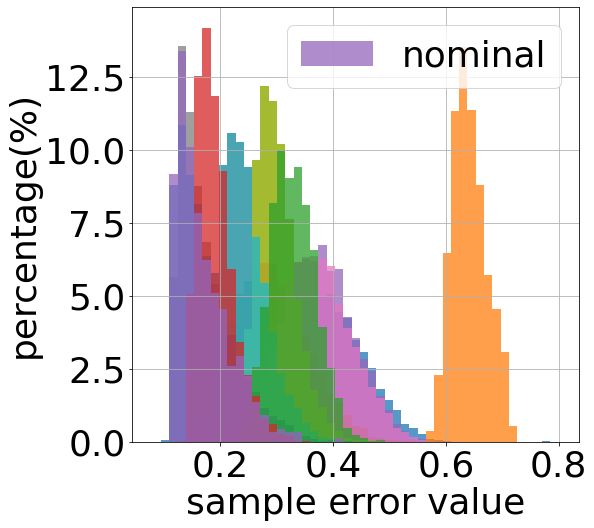}}
\subfigure[Re-trained]{\label{fig:b}\includegraphics[width=40.5mm]{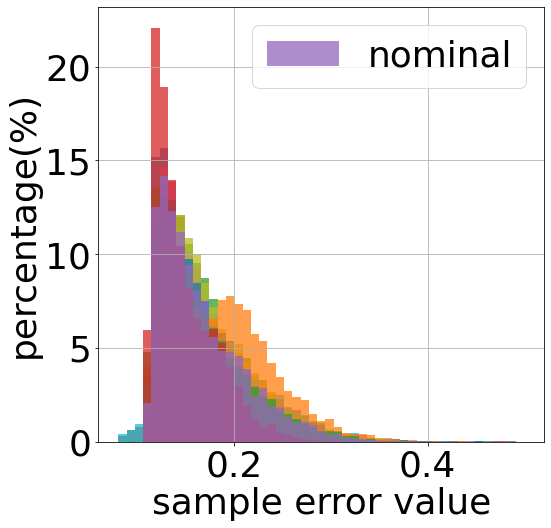}}
\vspace*{-3mm}
\caption{The distribution of sample $L_2$ prediction error of (a) the  pre-trained GNN on randomly perturbed grids and (b) after fast re-training. Each color indicates a new topology.}
\vspace*{-3mm}
\label{fig:topo_adap}
\end{figure}

{\bf Topology adaptivity:} We have further tested the 118-dc OPF case to validate the topology adaptivity of proposed GNN-based models. Specifically, after obtaining the trained GNN model for the nominal topology, we randomly pick at most two lines to disconnect and test the pre-trained GNN models on this new topology. Figure \ref{fig:topo_adap}(a) shows that pre-trained GNN models attain satisfactory prediction performance for some new topologies. In addition, we have implemented a post-processing step by using the pre-trained GNNs as warm start for re-training under each new topology. The post-processing step attains very fast convergence with just $3-5$ epochs, and high prediction performance as shown in Figure \ref{fig:topo_adap}(b). This result  demonstrates that GNN models are promising in adapting to real-time power grid topology, and points to an exciting future research direction.

\section{Conclusion and Future Work}
This paper proposes a new GNN-based approach for predicting the electricity market prices in order to support the efficient and reliable operations of low-carbon electric grids. Different from earlier learning-for-OPF approaches, our proposed method innovatively incorporates electricity prices' locality property  and physics-based regularization term to the design of topology-aware GNN models. Reduced model complexity and topology adaptivity are attained by the GNN-based price prediction. Numerical tests have demonstrated the efficiency and adaptivity of our price prediction method. Interesting future research directions open up on the formal investigation of topology adaptivity and other transfer learning aspects, as well as the extension to general optimal resource allocation problems in networked systems. 


\bibliography{ref.bib}

\begin{thebibliography}{8}
\providecommand{\natexlab}[1]{#1}
\providecommand{\url}[1]{\texttt{#1}}
\expandafter\ifx\csname urlstyle\endcsname\relax
  \providecommand{\doi}[1]{doi: #1}\else
  \providecommand{\doi}{doi: \begingroup \urlstyle{rm}\Url}\fi

\bibitem[Author(2021)]{anonymous}
Author, N.~N.
\newblock Suppressed for anonymity, 2021.

\bibitem[Duda et~al.(2000)Duda, Hart, and Stork]{DudaHart2nd}
Duda, R.~O., Hart, P.~E., and Stork, D.~G.
\newblock \emph{Pattern Classification}.
\newblock John Wiley and Sons, 2nd edition, 2000.

\bibitem[Kearns(1989)]{kearns89}
Kearns, M.~J.
\newblock \emph{Computational Complexity of Machine Learning}.
\newblock PhD thesis, Department of Computer Science, Harvard University, 1989.

\bibitem[Langley(2000)]{langley00}
Langley, P.
\newblock Crafting papers on machine learning.
\newblock In Langley, P. (ed.), \emph{Proceedings of the 17th International
  Conference on Machine Learning (ICML 2000)}, pp.\  1207--1216, Stanford, CA,
  2000. Morgan Kaufmann.

\bibitem[Michalski et~al.(1983)Michalski, Carbonell, and
  Mitchell]{MachineLearningI}
Michalski, R.~S., Carbonell, J.~G., and Mitchell, T.~M. (eds.).
\newblock \emph{Machine Learning: An Artificial Intelligence Approach, Vol. I}.
\newblock Tioga, Palo Alto, CA, 1983.

\bibitem[Mitchell(1980)]{mitchell80}
Mitchell, T.~M.
\newblock The need for biases in learning generalizations.
\newblock Technical report, Computer Science Department, Rutgers University,
  New Brunswick, MA, 1980.

\bibitem[Newell \& Rosenbloom(1981)Newell and Rosenbloom]{Newell81}
Newell, A. and Rosenbloom, P.~S.
\newblock Mechanisms of skill acquisition and the law of practice.
\newblock In Anderson, J.~R. (ed.), \emph{Cognitive Skills and Their
  Acquisition}, chapter~1, pp.\  1--51. Lawrence Erlbaum Associates, Inc.,
  Hillsdale, NJ, 1981.

\bibitem[Samuel(1959)]{Samuel59}
Samuel, A.~L.
\newblock Some studies in machine learning using the game of checkers.
\newblock \emph{IBM Journal of Research and Development}, 3\penalty0
  (3):\penalty0 211--229, 1959.

\end{thebibliography}


\begin{thebibliography}{25}
\providecommand{\natexlab}[1]{#1}
\providecommand{\url}[1]{\texttt{#1}}
\expandafter\ifx\csname urlstyle\endcsname\relax
  \providecommand{\doi}[1]{doi: #1}\else
  \providecommand{\doi}{doi: \begingroup \urlstyle{rm}\Url}\fi

\bibitem[Babaeinejadsarookolaee et~al.(2019)Babaeinejadsarookolaee, Birchfield,
  Christie, Coffrin, DeMarco, Diao, Ferris, Fliscounakis, Greene, Huang,
  et~al.]{babaeinejadsarookolaee2019power}
Babaeinejadsarookolaee, S., Birchfield, A., Christie, R.~D., Coffrin, C.,
  DeMarco, C., Diao, R., Ferris, M., Fliscounakis, S., Greene, S., Huang, R.,
  et~al.
\newblock The power grid library for benchmarking ac optimal power flow
  algorithms.
\newblock \emph{arXiv preprint arXiv:1908.02788}, 2019.

\bibitem[Baker(2019)]{baker2019learning}
Baker, K.
\newblock Learning warm-start points for ac optimal power flow.
\newblock In \emph{2019 IEEE 29th International Workshop on Machine Learning
  for Signal Processing (MLSP)}, pp.\  1--6. IEEE, 2019.

\bibitem[Birchfield et~al.(2016)Birchfield, Xu, Gegner, Shetye, and
  Overbye]{birchfield2016grid}
Birchfield, A.~B., Xu, T., Gegner, K.~M., Shetye, K.~S., and Overbye, T.~J.
\newblock Grid structural characteristics as validation criteria for synthetic
  networks.
\newblock \emph{IEEE Transactions on power systems}, 32\penalty0 (4):\penalty0
  3258--3265, 2016.

\bibitem[Boyd et~al.(2004)Boyd, Boyd, and Vandenberghe]{boyd2004convex}
Boyd, S., Boyd, S.~P., and Vandenberghe, L.
\newblock \emph{Convex optimization, Ch. 3 \& Ch. 5.5}.
\newblock Cambridge university press, 2004.

\bibitem[Cain et~al.(2012)Cain, O’neill, Castillo, et~al.]{cain2012history}
Cain, M.~B., O’neill, R.~P., Castillo, A., et~al.
\newblock History of optimal power flow and formulations.
\newblock \emph{Federal Energy Regulatory Commission}, 1:\penalty0 1--36, 2012.

\bibitem[Chen \& Zhang(2020)Chen and Zhang]{chen2020learning}
Chen, Y. and Zhang, B.
\newblock Learning to solve network flow problems via neural decoding.
\newblock \emph{arXiv preprint arXiv:2002.04091}, 2020.

\bibitem[Deka \& Misra(2019)Deka and Misra]{deka2019learning}
Deka, D. and Misra, S.
\newblock Learning for dc-opf: Classifying active sets using neural nets.
\newblock In \emph{2019 IEEE Milan PowerTech}, pp.\  1--6. IEEE, 2019.

\bibitem[Gama et~al.(2020)Gama, Isufi, Leus, and Ribeiro]{gama2020graphs}
Gama, F., Isufi, E., Leus, G., and Ribeiro, A.
\newblock Graphs, convolutions, and neural networks: From graph filters to
  graph neural networks.
\newblock \emph{IEEE Signal Processing Magazine}, 37\penalty0 (6):\penalty0
  128--138, 2020.

\bibitem[Garcia(2019)]{garcia2019non}
Garcia, M.~J.
\newblock \emph{Non-convex myopic electricity markets: the AC transmission
  network and interdependent reserve types, Ch. 5 \& Ch. 6}.
\newblock PhD thesis, 2019.

\bibitem[Garg et~al.(2020)Garg, Jegelka, and Jaakkola]{garg2020generalization}
Garg, V., Jegelka, S., and Jaakkola, T.
\newblock Generalization and representational limits of graph neural networks.
\newblock In \emph{International Conference on Machine Learning}, pp.\
  3419--3430. PMLR, 2020.

\bibitem[Geng \& Xie(2016)Geng and Xie]{geng2016learning}
Geng, X. and Xie, L.
\newblock Learning the lmp-load coupling from data: A support vector machine
  based approach.
\newblock \emph{IEEE Transactions on Power Systems}, 32\penalty0 (2):\penalty0
  1127--1138, 2016.

\bibitem[Guha et~al.(2019)Guha, Wang, Wytock, and Majumdar]{guha2019machine}
Guha, N., Wang, Z., Wytock, M., and Majumdar, A.
\newblock Machine learning for ac optimal power flow.
\newblock In \emph{n Climate Change Workshop at The Thirty-sixth International
  Conference on Machine Learning (ICML)}, 2019.

\bibitem[Isufi et~al.(2020)Isufi, Gama, and Ribeiro]{isufi2020edgenets}
Isufi, E., Gama, F., and Ribeiro, A.
\newblock Edgenets: Edge varying graph neural networks.
\newblock \emph{arXiv preprint arXiv:2001.07620}, 2020.

\bibitem[Jia et~al.(2013)Jia, Kim, Thomas, and Tong]{jia2013impact}
Jia, L., Kim, J., Thomas, R.~J., and Tong, L.
\newblock Impact of data quality on real-time locational marginal price.
\newblock \emph{IEEE Transactions on Power Systems}, 29\penalty0 (2):\penalty0
  627--636, 2013.

\bibitem[Kipf \& Welling(2016)Kipf and Welling]{kipf2016semi}
Kipf, T.~N. and Welling, M.
\newblock Semi-supervised classification with graph convolutional networks.
\newblock \emph{arXiv preprint arXiv:1609.02907}, 2016.

\bibitem[Ma \& Tang(2020)Ma and Tang]{ma2020deep}
Ma, Y. and Tang, J.
\newblock \emph{Deep Learning on Graphs}.
\newblock Cambridge University Press, 2020.

\bibitem[Misra et~al.(2018)Misra, Roald, and Ng]{misra2018learning}
Misra, S., Roald, L., and Ng, Y.
\newblock Learning for constrained optimization: Identifying optimal active
  constraint sets.
\newblock \emph{arXiv preprint arXiv:1802.09639}, 2018.

\bibitem[Owerko et~al.(2020)Owerko, Gama, and Ribeiro]{owerko2019optimal}
Owerko, D., Gama, F., and Ribeiro, A.
\newblock Optimal power flow using graph neural networks.
\newblock In \emph{ICASSP 2020 - 2020 IEEE International Conference on
  Acoustics, Speech and Signal Processing (ICASSP)}, pp.\  5930--5934, 2020.
\newblock \doi{10.1109/ICASSP40776.2020.9053140}.

\bibitem[Pan et~al.(2019)Pan, Zhao, and Chen]{pan2019deepopf}
Pan, X., Zhao, T., and Chen, M.
\newblock Deepopf: Deep neural network for dc optimal power flow.
\newblock In \emph{2019 IEEE International Conference on Communications,
  Control, and Computing Technologies for Smart Grids (SmartGridComm)}, pp.\
  1--6. IEEE, 2019.

\bibitem[Price \& Goodin(2011)Price and Goodin]{price2011reduced}
Price, J.~E. and Goodin, J.
\newblock Reduced network modeling of wecc as a market design prototype.
\newblock In \emph{2011 IEEE Power and Energy Society General Meeting}, pp.\
  1--6. IEEE, 2011.

\bibitem[Ramakrishna \& Scaglione(2021)Ramakrishna and
  Scaglione]{scaglione2021gsp}
Ramakrishna, R. and Scaglione, A.
\newblock Grid-graph signal processing (grid-gsp): A graph signal processing
  framework for the power grid.
\newblock \emph{IEEE Transactions on Signal Processing}, pp.\  1--1, 2021.
\newblock \doi{10.1109/TSP.2021.3075145}.

\bibitem[Wood et~al.(2013)Wood, Wollenberg, and Shebl{\'e}]{wood2013power}
Wood, A.~J., Wollenberg, B.~F., and Shebl{\'e}, G.~B.
\newblock \emph{Power generation, operation, and control, Sec. 3.10}.
\newblock John Wiley \& Sons, 2013.

\bibitem[Zamzam \& Baker(2020)Zamzam and Baker]{zamzam2020learning}
Zamzam, A.~S. and Baker, K.
\newblock Learning optimal solutions for extremely fast ac optimal power flow.
\newblock In \emph{2020 IEEE International Conference on Communications,
  Control, and Computing Technologies for Smart Grids (SmartGridComm)}, pp.\
  1--6. IEEE, 2020.

\bibitem[Zamzam \& Sidiropoulos(2020)Zamzam and
  Sidiropoulos]{zamzam2020physics}
Zamzam, A.~S. and Sidiropoulos, N.~D.
\newblock Physics-aware neural networks for distribution system state
  estimation.
\newblock \emph{IEEE Transactions on Power Systems}, 2020.

\bibitem[{Zimmerman} et~al.(2011){Zimmerman}, {Murillo-Sánchez}, and
  {Thomas}]{matpower}
{Zimmerman}, R.~D., {Murillo-Sánchez}, C.~E., and {Thomas}, R.~J.
\newblock Matpower: Steady-state operations, planning, and analysis tools for
  power systems research and education.
\newblock \emph{IEEE Transactions on Power Systems}, 26\penalty0 (1):\penalty0
  12--19, 2011.

\end{thebibliography}
\bibliographystyle{icml2021}





\end{document}


\twocolumn[
\icmltitle{Graph Neural Networks for Learning Real-Time
    Prices in Electricity Market}



\icmlsetsymbol{equal}{*}

\begin{icmlauthorlist}
\icmlauthor{Shaohui Liu}{to}
\icmlauthor{Chengyang Wu}{to}
\icmlauthor{Hao Zhu}{to}
\end{icmlauthorlist}

\icmlaffiliation{to}{Department of Electrical and Computer Engineering, University of Texas at Austin, Austin, United States}

\icmlcorrespondingauthor{Shaohui Liu}{shaohui.liu@utexas.edu}
\icmlcorrespondingauthor{Hao Zhu}{haozhu@utexas.edu}

\icmlkeywords{graph neural network, optimal power flow, real-time electricity pricing, transfer learning}

\vskip 0.3in
]






\section{Supplementary Materials}

In the submitted manuscript, comparisons of $L_2$ errors along with feasiblity ratios of models are illustrated as scatter plots. Here we provide the detailed statistical values in Table \ref{table1} for reference. Note that for the 2383dc case, although the feasibility results are all $99.9\%$, models with feeasibility ragularization are in general with 2 orders of higher feasibility for all cases.

In addition, we also summarize model parameters, occupied memories, training times, and number of training epochs of all models on different systems in Table \ref{table2}.

\begin{table}[h]
\centering
\begin{tabular}{lllll}
\hline
Method & Metric & 118ac & 118dc & 2383dc \\ \hline
GNN+FR & $L_2$ & 5.3e-2 & 6.4e-2 & 1-6.8e-2 \\
& Feas. & 99.3\% & 99.4\% & $99.9\%$ \\ \hline
GNN & $L_2$ & 5.2e-2 & 6.1e-2 & 6.8e-2 \\
& Feas. & 99.1\% & 99.0\% & $99.9\%$ \\ \hline
FCNN+FR & $L_2$ & 4.5e-2 & 4.5e-2 & 4.5e-2 \\
& Feas. & 98.9\% & 90.7\% & $99.9\%$ \\ \hline
FCNN & $L_2$ & 4.4e-2 & 4.4e-2 & 4.5e-2 \\
& Feas. & 98.8\% & 90.7\% & $99.9\%$ \\ \hline
GiDNN+FR & $L_2$ & 4.4e-2 & 5.5e-2 & 5.6e-2 \\
& Feas. & 99.4\% & 99.4\% & $99.9\%$ \\ \hline
GiDNN & $L_2$ & 5.3e-2 & 5.0e-2 & 5.6e-2 \\
& Feas. & 99.3\% & 98.5\% & $99.9\%$ \\ \hline
\end{tabular}
\caption{Performance of GNN, FCNN, and GiDNN in predicting price and solution feasibility}
\label{table1}
\end{table}

\newpage
\begin{table}[tb!]
\centering
\begin{tabular}{lllll}
\hline
Method & Metric & 118ac & 118dc & 2383dc \\ \hline
GNN+FR & Parameters & 351K & 351K & 142M \\
& Memory & 1365MB & 1369MB & 1919MB \\
& Time & 29s & 95s & 103s \\
& Epoch & 49 & 69 & 54 \\ \hline
GNN & Parameters & 351K & 351K & 142M \\
& Memory & 935MB & 939MB & 1919MB \\
& Time & 35s & 160s & 159s \\
& Epoch & 114 & 134 & 64 \\ \hline
FCNN+FR & Parameters & 5.8M & 5.8M & 284M \\
& Memory & 953MB & 953MB & 2295MB \\
& Time & 24s & 12s & 166s \\
& Epoch & 120 & 30 & 35 \\ \hline
FCNN & Parameters & 5.8M & 5.8M & 284M \\
& Memory & 953MB & 953MB & 2295MB \\
& Time & 22s & 33s & 146s \\
& Epoch & 155 & 90 & 40 \\ \hline
PruneGNN+FR & Parameters & 9.7M & 9.4M & 210M \\
& Memory & 1019MB & 1023MB & 2539MB \\
& Time & 27s & 36s & 112s \\
& Epoch & 79 & 94 & 39 \\ \hline
PruneGNN & Parameters & 9.7M & 9.4M & 210M \\
& Memory & 1019MB & 1023MB & 2539MB \\
& Time & 18s & 58s & 92s \\
& Epoch & 94 & 104 & 49 \\ \hline
\end{tabular}
\caption{Model complexity in number of parameters and memory, and training information in time and number of epochs of GNN, FCNN, and GiDNN on different datasets}
\label{table2}
\end{table}